\documentclass{article} % For LaTeX2e
\usepackage{iclr2016_conference,times}
\usepackage{hyperref}
\usepackage{url}
\usepackage{graphicx} % more modern
\usepackage{amsthm}
\usepackage{amsmath}
\usepackage{amsfonts}
\usepackage{bbm}
\usepackage{titlesec}
\usepackage{caption}
\usepackage{subcaption}
\usepackage{wrapfig}
\usepackage{appendix}
\usepackage{booktabs}

\newcommand*\samethanks[1][\value{footnote}]{\footnotemark[#1]}

\title{Learning to Diagnose with LSTM Recurrent Neural Networks}

\author{Zachary C. Lipton \thanks{Equal contributions}\ \thanks{Author website: \texttt{http://zacklipton.com} } \\
Department of Computer Science and Engineering\\
University of California, San Diego\\
La Jolla, CA 92093, USA \\
\url{zlipton@cs.ucsd.edu} \\
\And
David C. Kale \samethanks[1]\ \thanks{Author website: \url{http://www-scf.usc.edu/\textasciitilde dkale/} } \\
Department of Computer Science \textcolor{white}{PUSH OVER TO LEFT}\\
University of Southern California \\
Los Angeles, CA 90089 \\
\texttt{dkale@usc.edu} \\
\And
Charles Elkan \\
Department of Computer Science and Engineering \\
University of California, San Diego\\
La Jolla, CA 92093, USA \\
\texttt{elkan@cs.ucsd.edu} \\
\And
Randall Wetzel \\
Laura P. and Leland K. Whittier Virtual PICU \\
Children's Hospital Los Angeles \\
Los Angeles, CA 90027 \\
\texttt{rwetzel@chla.usc.edu} \\
}

\iclrfinalcopy % Uncomment for camera-ready version

\begin{document}

\maketitle

\begin{abstract}
Clinical medical data, especially in the intensive care unit (ICU),
consist of multivariate time series of observations.
For each patient visit (or \emph{episode}), 
sensor data and lab test results 
are recorded in the patient's Electronic Health Record (EHR).
While potentially containing a wealth of insights, 
the data is difficult to mine effectively,
owing to varying length, irregular sampling and missing data.
Recurrent Neural Networks (RNNs), 
particularly those using Long Short-Term Memory (LSTM) hidden units,
are powerful and increasingly popular models for learning from sequence data.
They effectively model varying length sequences 
and capture long range dependencies.
We present the first study to empirically evaluate the ability of LSTMs 
to recognize patterns in multivariate time series of clinical measurements.
Specifically, we consider multilabel classification of diagnoses,
training a model to classify $128$ diagnoses 
given $13$ frequently but irregularly sampled clinical measurements.
First, we establish the effectiveness of a simple LSTM network for modeling clinical data.
Then we demonstrate a straightforward and effective training strategy 
in which we replicate targets at each sequence step.
Trained only on raw time series, 
our models outperform several strong baselines,
including a multilayer perceptron 
trained on hand-engineered features.
\end{abstract}

\section{Introduction}
Time series data comprised of clinical measurements,
as recorded by caregivers in the pediatric intensive care unit (PICU), 
constitute an abundant and largely untapped source of medical insights.
Potential uses of such data 
include classifying diagnoses accurately, predicting length of stay, 
predicting future illness, and predicting mortality.
However, besides the difficulty of acquiring data, 
several obstacles stymie machine learning research with clinical time series.
Episodes vary in length, 
with stays ranging from just a few hours to multiple months.
Observations, which include sensor data, vital signs, lab test results, and subjective assessments, 
are sampled irregularly and plagued by missing values \citep{marlin:ihi2012}.
Additionally, long-term time dependencies complicate learning with many algorithms.
Lab results that, taken together, might imply a particular diagnosis 
may be separated by days or weeks.
Long delays often separate onset of 
disease from the appearance of symptoms.
For example, symptoms of acute respiratory distress syndrome 
may not appear until $24$-$48$ hours after lung injury \citep{mason2010murray}, 
while symptoms of an asthma attack may present shortly after admission but change or disappear following treatment.

Recurrent Neural Networks (RNNs),
in particular those based on Long Short-Term Memory (LSTM) \citep{hochreiter1997long}, 
model varying-length sequential data,
achieving state-of-the-art results for problems 
spanning natural language processing, image captioning, handwriting recognition,
and genomic analysis \citep{auli2013joint, sutskever2014sequence, vinyals2014show, karpathy2014deep, liwicki2007novel, graves2009novel, pollastri2002improving, vohradsky2001neural, xu2007inference}. 
LSTMs can capture long range dependencies and nonlinear dynamics. 
Some sequence models, such as Markov models, conditional random fields, and Kalman filters, deal with sequential data 
but are ill-equipped to learn long-range dependencies.
Other models require domain knowledge or feature engineering, 
offering less chance for serendipitous discovery. 
In contrast, neural networks learn representations and can discover unforeseen structure.

This paper presents the first empirical study 
using LSTMs to classify diagnoses 
given multivariate PICU time series.
Specifically, we formulate the problem as multilabel classification, 
since diagnoses are not mutually exclusive.
Our examples are clinical episodes,
each consisting of $13$ frequently but irregularly sampled time series of clinical measurements,
including body temperature, heart rate, 
diastolic and systolic blood pressure, 
and blood glucose, among others. 
Associated with each patient are a subset of $429$ diagnosis codes.
As some are rare, we focus on the $128$ most common codes, 
classifying each episode with one or more diagnoses.

Because LSTMs have never been used in this setting,
we first verify their utility and compare their performance to a set of strong baselines,
including both a linear classifier and a MultiLayer Perceptron (MLP).
We train the baselines on both a fixed window and hand-engineered features.
We then test a straightforward \emph{target replication} strategy 
for recurrent neural networks, 
inspired by the \emph{deep supervision} technique of \cite{lee2014deeply} 
for training convolutional neural networks.
We compose our optimization objective 
as a convex combination of the loss at the final sequence step 
and the mean of the losses over \emph{all} sequence steps.
Additionally, we evaluate the efficacy 
of using additional information in the patient's chart as auxiliary outputs, 
a technique previously used with feedforward nets \citep{caruana1996using}, showing that it reduces overfitting.
Finally, we apply dropout to non-recurrent connections,
which improves the performance further.
LSTMs with target replication and dropout
surpass the performance of the best baseline, 
namely an MLP trained on hand-engineered features,
even though the LSTM has access only to raw time series.

\section{Related Work}
Our research sits at the intersection of LSTMs, 
medical informatics, and multilabel classification, 
three  mature fields, 
each with a long history and rich body of research.
While we cannot do justice to all three,
we highlight the most relevant works below.

\subsection{LSTM RNNs}
LSTMs were originally introduced in
\cite{hochreiter1997long}, 
following a long line of research into RNNs for sequence learning. 
Notable earlier work includes \citet{rumelhart1985learning},
which introduced backpropagation through time,
and \citet{elman1990finding},
which successfully trained RNNs 
to perform supervised machine learning tasks 
with sequential inputs and outputs.
The design of modern LSTM memory cells 
has remained close to the original,
with the commonly used addition of forget gates \citep{gers2000learning} (which we use), 
and peep-hole connections \citep{gers2000count} (which we do not use).
The connectivity pattern among multiple LSTM layers in our models follows the architecture described by \cite{graves2013generating}.
\citet{pascanu2013construct} explores other mechanisms by which
an RNN could be made \emph{deep}.
Surveys of the literature include \citet{graves2012supervised}, 
a thorough dissertation on sequence labeling with RNNs,
\citet{de2015survey}, which surveys natural language applications, 
and \cite{lipton2015critical}, which provides a broad overview 
of RNNs for sequence learning, focusing on modern applications.

\subsection{Neural Networks for Medical Data}
Neural networks have been applied to medical problems and data 
for at least 20 years \citep{caruana1996using,baxt1995lancet}, 
although we know of no work on applying LSTMs to multivariate clinical time series 
of the type we analyze here. 
Several papers have applied RNNs to physiologic signals, 
including electrocardiograms \citep{silipo1998artificial, amari1998adaptive, ubeyli2009combining} 
and glucose measurements \citep{tresp:nips1997}. 
RNNs have also been used for prediction problems in genomics 
\citep{pollastri2002improving, xu2007inference, vohradsky2001neural}. 
Multiple recent papers apply modern deep learning techniques (but not RNNs) 
to modeling psychological conditions \citep{dabek2015neural}, 
head injuries \citep{rughani2010use}, and Parkinson's disease \citep{hammerla2015pd}.
Recently, feedforward networks have been 
applied to medical time series in sliding window fashion
to classify cases of gout, leukemia \citep{lasko:plosone2013}, and critical illness \citep{che:kdd2015}.  

\subsection{Neural Networks for Multilabel Classification}
Only a few published papers apply LSTMs to multilabel classification tasks,
all of which, to our knowledge, are outside of the medical context.
\cite{liu2014bach} formulates music composition as a multilabel classification task, 
using sigmoidal output units. 
Most recently, \citet{yeung2015every}
uses LSTM networks with multilabel outputs to 
recognize actions in videos.
While we could not locate any published papers using
LSTMs for multilabel classification in the medical domain,
several papers use feedforward nets for this task.
One of the earliest papers to investigate multi-task neural networks 
modeled risk in pneumonia patients \citep{caruana1996using}. 
More recently, \citet{che:kdd2015} formulated diagnosis 
as multilabel classification 
using a sliding window multilayer perceptron.

\subsection{Machine Learning for Clinical Time Series}
Neural network methodology aside, 
a growing body of research applies machine learning 
to temporal clinical data for tasks including artifact removal \citep{aleks2009probabilistic,quinn2009factorial},
early detection and prediction \citep{stanculescu2014autoregressive,henry2015targeted},
and clustering and subtyping \citep{marlin:ihi2012,schulam2015clustering}.
Many recent papers use models with latent factors 
to capture nonlinear dynamics in clinical time series 
and to discover meaningful representations of health and illness. 
Gaussian processes are popular because they can directly handle irregular sampling 
and encode prior knowledge via choice of covariance functions 
between time steps and across variables \citep{marlin:ihi2012,ghassemi2015multivariate}. 
\citet{saria2010learning} combined a hierarchical dirichlet process 
with autoregressive models to infer latent disease ``topics'' 
in the heart rate signals of premature babies. 
\citet{quinn2009factorial} used linear dynamical systems 
with latent switching variables 
to model physiologic events like bradycardias. 
Seeking \emph{deeper} models,
\cite{stanculescuhierarchical} proposed a second ``layer'' of latent factors 
to capture correlations between latent states.

\subsection{Target Replication}
In this work, we make the task of classifying entire sequences easier
by replicating targets at every time step,
inspired by \cite{lee2014deeply}, 
who place an optimization objective after each layer in convolutional neural network.
While they have a separate set of weights to learn each intermediate objective,
our model is simpler owing to the weight tying in recurrent nets, having only one set of output weights.
Additionally, unlike \citet{lee2014deeply}, we place targets at each time step, but not following each layer between input and output in the LSTM.
After finishing this manuscript, 
we learned that target replication strategies similar to ours
have also been developed by \citet{ng2015beyond} and \citet{dai2015semi} 
for the tasks of video classification 
and character-level document classification respectively.
\citet{ng2015beyond}
linearly scale the importance of each intermediate target, 
emphasizing performance at later sequence steps 
over those in the beginning of the clip.
\citet{dai2015semi} also use a target replication strategy 
with linearly increasing weight 
for character-level document classification, 
showing significant improvements in accuracy. 
They call this technique \emph{linear gain}.

\subsection{Regularizing Recurrent Neural Networks}
Given the complexity of our models and modest scale of our data,
regularization, including judicious use of dropout, is crucial to our performance. 
Several prior works use dropout to regularize RNNs.
\citet{pham2014dropout}, \citet{zaremba2014recurrent}, and \citet{dai2015semi}
all describe an application of dropout to only the non-recurrent weights of a network. 
The former two papers establish the method and apply it to tasks with sequential outputs, including handwriting recognition, image captioning, and machine translation.
The setting studied by \citet{dai2015semi} most closely resembles ours as the authors apply it to the task of applying static labels to varying length sequences.

\subsection{Key Differences}
Our experiments show that LSTMs can accurately classify  multivariate time series of clinical measurements, 
a topic not addressed in any prior work.
Additionally, while some papers use LSTMs for multilabel classification,
ours is the first to address this problem in the medical context. Moreover, for multilabel classification of sequential clinical data
with fixed length output vectors, 
this paper is the first, to our knowledge, 
to demonstrate the efficacy of a target replication strategy,
achieving both faster training and better generalization.

\section{Data Description}
Our experiments use a collection of anonymized clinical time series 
extracted from the EHR system at Children's Hospital LA \citep{marlin:ihi2012,che:kdd2015} 
as part of an IRB-approved study. 
The data consists of $10,401$ PICU episodes, 
each a multivariate time series of 13 variables: 
diastolic and systolic blood pressure, peripheral capillary refill rate,
end-tidal CO$_2$, fraction of inspired O$_2$,
Glascow coma scale, blood glucose, heart rate, pH, 
respiratory rate, blood oxygen saturation,
body temperature, and urine output. Episodes vary in length from 12 hours to several months.

Each example consists of irregularly sampled multivariate time series 
with both missing values and, occasionally, missing variables.
We resample all time series to an hourly rate, 
taking the mean measurement within each one hour window. 
We use forward- and back-filling to fill gaps created 
by the window-based resampling.
When a single variable's time series is missing entirely, 
we impute a clinically normal value as defined by domain experts.
These procedures make reasonable assumptions about clinical practice: 
many variables are recorded at rates proportional to how quickly they change, and when a variable is absent, 
it is often because clinicians believed it to be normal 
and chose not to measure it.
Nonetheless, these procedures are not appropriate in all settings. 
Back-filling, for example, passes information from the future backwards. 
This is acceptable for classifying entire episodes (as we do)
but not for forecasting. 
Finally, we rescale all variables to $[0,1]$, 
using ranges defined by clinical experts.
In addition, we use published tables of normal values 
from large population studies 
to correct for differences in heart rate, respiratory rate, \citep{fleming2011lancet} and blood pressure \citeyearpar[NHBPEP Working Group][]{nhbp2004report} due to age and gender.

Each episode is associated with zero or more diagnostic codes 
from an in-house taxonomy used for research and billing, 
similar to the \textit{Ninth Revision of the International Classification of Diseases} (ICD-9) codes \citep{world2004international}. 
The dataset contains $429$ distinct labels indicating a variety of conditions, 
such as acute respiratory distress, congestive heart failure, seizures, 
renal failure, and sepsis. 
Because many of the diagnoses are rare,
we focus on the most common $128$,
each of which occurs more than $50$ times in the data. These diagnostic codes are recorded by attending physicians during or shortly after each patient episode and subject to limited review afterwards.

Because the diagnostic codes were assigned by clinicians, 
our experiments represent a comparison of an LSTM-based diagnostic system to human experts. 
We note that an attending physician has access to much more data about each patient than our LSTM does, 
including additional tests, medications, and treatments. 
Additionally, the physician can access a full medical history including free-text notes,
can make visual and physical inspections of the patient, and can ask questions.
A more fair comparison might require asking additional clinical experts to assign diagnoses given access only to the 13 time series available to our models. 
However, this would be prohibitively expensive, even for just the $1000$ examples, and difficult to justify to our medical collaborators, as this annotation would provide no immediate benefit to patients. Such a study will prove more feasible in the future when this line of research has matured.

\section{Methods}
In this work, we are interested in recognizing diagnoses and, more broadly, 
the observable physiologic characteristics of patients, 
a task generally termed \textit{phenotyping} \citep{oellrich2015phenotyping}.
We cast the problem of phenotyping clinical time series as multilabel classification. Given a series of observations $\boldsymbol{x}^{(1)},...,\boldsymbol{x}^{(T)}$,
we learn a classifier to generate hypotheses $\boldsymbol{\hat{y}}$
of the true labels $\boldsymbol{y}$.
Here, $t$ indexes sequence steps, and for any example, 
$T$ stands for the length of the sequence.
Our proposed LSTM RNN uses memory cells with forget gates \citep{gers2000learning} 
but without peephole connections \citep{gers2003learning}. 
As output, we use a fully connected layer atop the highest LSTM layer 
followed by an element-wise sigmoid activation function, 
because our problem is multilabel.
We use \emph{log loss} as the loss function at each output.

The following equations give the update for a layer of memory cells $\boldsymbol{h}_l^{(t)}$ where $\boldsymbol{h}_{l-1}^{(t)}$ stands for the previous layer at the same sequence step (a previous LSTM layer or the input $\boldsymbol{x}^{(t)}$) and $\boldsymbol{h}^{(t-1)}_{l}$ stands for the same layer at the previous sequence step:
$$ \boldsymbol{g}_l^{(t)} = \phi( W_l^{\mbox{gx}} \boldsymbol{h}^{(t)}_{l-1} +   W_l^{\mbox{gh}} \boldsymbol{h}^{(t-1)}_{l}  + \boldsymbol{b}_l^{\mbox{g}})$$
$$ \boldsymbol{i}_l^{(t)}  =    \sigma( W_l^{\mbox{ix}} \boldsymbol{h}^{(t)}_{l-1} + W_l^{\mbox{ih}} \boldsymbol{h}^{(t-1)}_{l} + \boldsymbol{b}_l^{\mbox{i}}) $$
$$ \boldsymbol{f}_l^{(t)}  =    \sigma( W_l^{\mbox{fx}} \boldsymbol{h}^{(t)}_{l-1} + W_l^{\mbox{fh}} \boldsymbol{h}^{(t-1)}_{l} + \boldsymbol{b}_l^{\mbox{f}}) $$
$$ \boldsymbol{o}_l^{(t)} =    \sigma( W_l^{\mbox{ox}} \boldsymbol{h}^{(t)}_{l-1} + W_l^{\mbox{oh}} \boldsymbol{h}^{(t-1)}_{l} + \boldsymbol{b}_l^{\mbox{o}}) $$
$$ \boldsymbol{s}_l^{(t)} = \boldsymbol{g}_l^{(t)} \odot \boldsymbol{i}_l^{(i)} + \boldsymbol{s}_l^{(t-1)} \odot \boldsymbol{f}_l^{(t)} $$
$$ \boldsymbol{h}^{(t)}_{l}  = \phi(\boldsymbol{s}_l^{(t)}) \odot \boldsymbol{o}_l^{(t)}. $$ 
In these equations, $\sigma$ stands for an element-wise application of the \textit{sigmoid} (\textit{logistic}) function, $\phi$ stands for an element-wise application of the $tanh$ function, and $\odot$ is the Hadamard (element-wise) product. The input, output, and forget gates are denoted by $\boldsymbol{i}$, $\boldsymbol{o}$, and $\boldsymbol{f}$ respectively, while $\boldsymbol{g}$ is the input node and has a \textit{tanh} activation.

\subsection{LSTM Architectures for Multilabel classification}
\begin{figure}[ht]
  \begin{center}
	\includegraphics[clip=true,trim=0 50 0 50,width=0.8\textwidth]{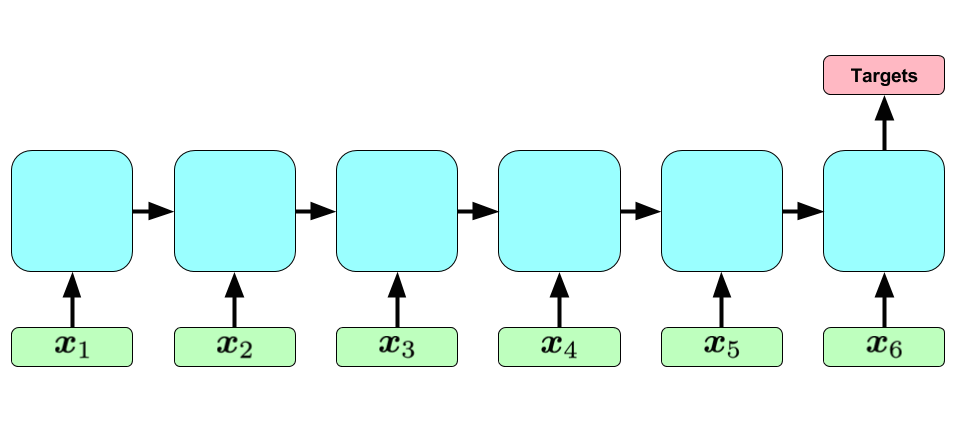}
  \end{center}
  \caption{A simple RNN model for multilabel classification. Green rectangles represent inputs. The recurrent hidden layers separating input and output are represented with a single blue rectangle. The red rectangle represents targets.}
\label{fig:simple-ml-rnn}
\end{figure}

We explore several recurrent neural network architectures 
for multilabel classification of time series.
The first and simplest (\autoref{fig:simple-ml-rnn}) passes over all inputs in chronological order,
generating outputs only at the final sequence step.
In this approach, we only have output $\hat{\boldsymbol{y}}$
at the final sequence step, 
at which our loss function is the average of the losses at each output node.
Thus the loss calculated at a single sequence step is
the average of \emph{log loss} calculated separately on each label.
$$\mbox{loss}(\boldsymbol{\hat{y}}, \boldsymbol{y}) = \frac{1}{|L|} \sum_{l=1}^{l=|L|} - (y_l \cdot \mbox{log}(\hat{y}_l) + (1 - y_l) \cdot \mbox{log}(1 - \hat{y}_l) ) .$$
\vspace{-0.2in}

\subsection{Sequential Target Replication}
\begin{figure}[ht]
  \begin{center}
	\includegraphics[clip=true,trim=0 50 0 45,width=0.8\textwidth]{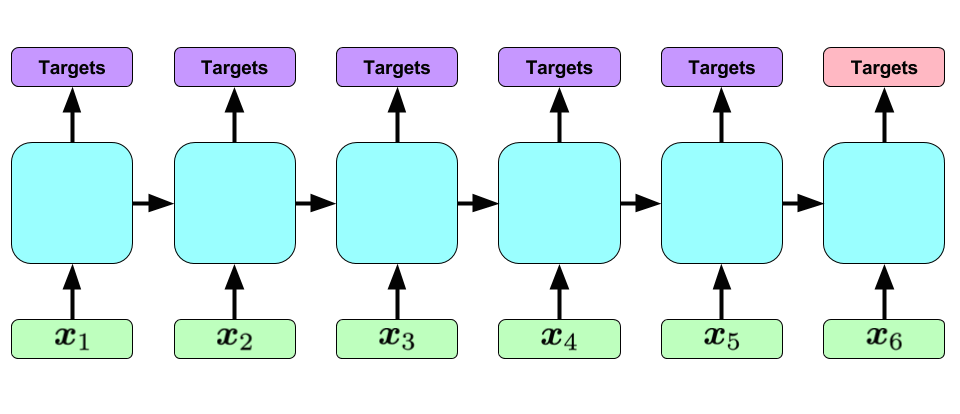}
  \end{center}
  \caption{An RNN classification model with \emph{target replication}. 
 The primary target (depicted in red) at the final step 
 is used at prediction time, 
 but during training, the model back-propagates errors 
 from the intermediate targets (purple) at every sequence step.}
\label{fig:icu-intermediate}
\end{figure}

One problem with the simple approach 
is that the network must learn to pass information 
across many sequence steps in order to affect the output.
We attack this problem by replicating our static targets at each sequence step (\autoref{fig:icu-intermediate}),
providing a local error signal at each step.
This approach is inspired by the deep supervision technique that \citet{lee2014deeply} apply to convolutional nets.
This technique is especially sensible in our case
because we expect the model to predict accurately even if the sequence were truncated by a small amount.
The approach differs from \citet{lee2014deeply}
because we use the same output weights to calculate $\boldsymbol{\hat{y}}^{(t)}$ for all $t$.
Further, we use this target replication to generate output at each sequence step, but not at each hidden layer. 

For the model with target replication, 
we generate an output $\hat{\boldsymbol{y}}^{(t)}$
at every sequence step.
Our loss is then a convex combination
of the final loss and the average of the losses over all steps:
$$\alpha \cdot \frac{1}{T} \sum_{t=1}^{T} \mbox{loss}(\boldsymbol{\hat{y}}^{(t)}, \boldsymbol{y}^{(t)})
+ (1-\alpha) \cdot \mbox{loss}(\boldsymbol{\hat{y}}^{(T)}, \boldsymbol{y}^{(T)})$$
where $T$ is the total number of sequence steps and $\alpha \in [0,1]$ is a hyper-parameter which determines the relative importance of hitting these intermediary targets.
At prediction time, we take only the output at the final step.
In our experiments, networks using target replication 
outperform those with a loss applied only at the final sequence step.

\subsection{Auxiliary Output Training}

\begin{figure}[ht]
  \begin{center}
	\includegraphics[clip=true,trim=0 30 0 30,width=0.8\textwidth]{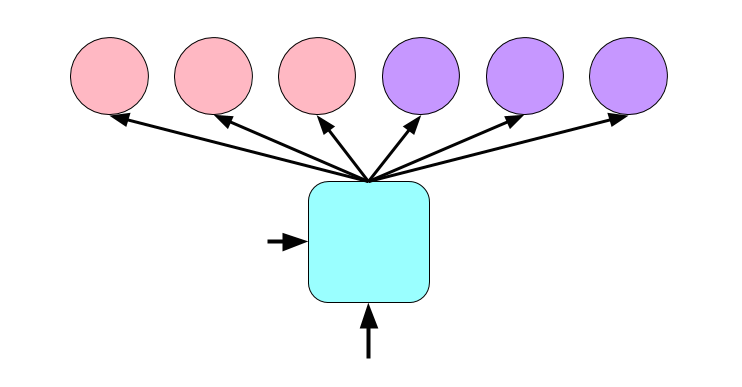}
  \end{center}
  \caption{Our dataset contains many labels. 
  For our task, a subset of $128$ are of interest (depicted in red). Our \emph{Auiliary Output} neural network makes use of extra labels as additional training targets (depicted in purple). At inference time we generate predictions for only the labels of interest.}
\label{fig:junkout}
\end{figure}

Recall that our initial data contained $429$ diagnostic labels
but that our task is to predict only $128$.
Given the well-documented successes 
of multitask learning with shared representations
and feedforward networks, 
we wish to train a stronger model by using the remaining $301$ labels 
or other information in the patient's chart, such as diagnostic categories, 
as auxiliary targets \citep{caruana1996using}.
These additional targets serve reduce overfitting 
as the model aims to minimize the loss on the labels of interest 
while also minimizing loss on the auxiliary targets (\autoref{fig:junkout}).

\subsection{Regularization}
Because we have only $10,401$ examples, overfitting is a considerable obstacle. 
Our experiments show that both target replication and auxiliary outputs improve performance and reduce overfitting. 
In addition to these less common techniques we deploy $\ell^2_2$ weight decay and dropout.
Following the example of \cite{zaremba2014recurrent} and \cite{pham2014dropout},
we apply dropout to the non-recurrent connections only.
We first compute each hidden layer's sequence of activations in the left-to-right direction and then apply dropout before computing the next layer's activations.
In our experiments, we find that dropout decreases overfitting, enabling us to double the size of each hidden layer.

\section{Experiments}
All models are trained on $80\%$ of the data 
and tested on $10\%$. 
The remaining $10\%$ is used as a validation set.
We train each LSTM for 100 epochs using stochastic gradient descent (SGD) with momentum.
To combat exploding gradients, 
we scale the norm of the gradient and
use $\ell_2^2$ weight decay of $10^{-6}$, 
both hyperparameters chosen using validation data.
Our final networks use $2$ hidden layers and either $64$ memory cells per layer with no dropout or $128$ cells per layer with dropout of $0.5$.
These architectures are also chosen based on validation performance. 
Throughout training, we save the model and compute three performance metrics (micro AUC, micro F1, and precision at 10) on the validation set for each epoch. 
We then test the model 
that scores best on at least two of the three validation metrics. 
To break ties, we choose the earlier epoch.

We evaluate a number of baselines
as well as LSTMs with various combinations of target replication (TR), dropout (DO), and auxiliary outputs (AO), using either the additional 301 diagnostic labels or 12 diagnostic categories. To explore the regularization effects of each strategy, we record and plot both training and validation performance after each epoch.
Additionally, we report performance of a target replication model (Linear Gain) that scales the weight of each intermediate target linearly as opposed our proposed approach.
Finally, to show that our LSTM learns a model complementary to the baselines, we evaluate an ensemble of the best LSTM with the best baseline.

\subsection{Multilabel Evaluation Methodology}
We report micro- and macro-averaged versions of Area Under the ROC Curve (AUC). 
By micro AUC, we mean a single AUC 
computed on flattened $\hat{Y}$ and $Y$ matrices,
whereas we calculate macro AUC by averaging each per-label AUC.
The blind classifier achieves $0.5$ macro AUC 
but can exceed $0.5$ on micro AUC 
by predicting labels in descending order by base rate.
Additionally, we report micro- and macro-averaged F1 score, computed in similar fashion to the respective micro and macro AUCs. 
F1 metrics require a thresholding strategy,
and here we select thresholds based upon validation set performance.
We refer to \cite{lipton2014optimal} for an analysis 
of the strengths and weaknesses of each type of multilabel F-score 
and a characterization of optimal thresholds.

Finally, we report \textit{precision at $10$}, 
which captures the fraction of true diagnoses 
among the model's top $10$ predictions, 
with a best possible score of $0.2281$ on the test split of this data set because there are on average $2.281$ diagnoses per patient.  
While F1 and AUC are both useful for determining the relative quality of a classifier's predictions, 
neither is tailored to a real-world application.
Thus, we consider a medically plausible use case 
to motivate this more interpretable metric:
generating a short list 
of the $10$ most probable diagnoses.
If we could create a high recall, moderate precision list of $10$ likely diagnoses, it could be a valuable hint-generation tool for differential diagnosis.
Testing for only the $10$ most probable conditions is much more realistic than testing for all conditions.

\subsection{Baseline Classifiers}
We provide results for a \textit{base rate} model 
that predicts diagnoses in descending order by incidence 
to provide a minimum performance baseline for micro-averaged metrics. 
We also report the performance of logistic regression,
which is widely used in clinical research. We train a separate classifier for each diagnosis 
but choose an overall $\ell_2^2$ penalty for all individual classifiers based on validation performance.
For a much stronger baseline, 
we train a multilabel MLP with 3 hidden layers of 300 hidden units each, rectified linear activations, and dropout of 0.5. All MLPs were trained for 1000 epochs, with hyperparameters chosen based on validation set performance. Each baseline is tested with two sets of inputs: 
raw time series and hand-engineered features. 
For raw time series, we use the first and last six hours. 
This provides classifiers with temporal information 
about changes in patient state from admission to discharge
within a fixed-size input, as required by all baselines. 
We find this works better than providing the first or last 12 hours alone.

Our hand-engineered features are inspired by those used in state-of-the-art severity of illness scores \citep{prism3}: for each variable, we compute the first and last measurements and their difference scaled by episode length, mean and standard deviation, median and quartiles, minimum and maximum, and slope of a line fit with least squares. These 143 features capture many of the indicators that clinicians look for in critical illness, including admission and discharge state, extremes, central tendencies, variability, and trends. They previously have been shown to be effective for these data \citep{marlin:ihi2012,che:kdd2015}. Our strongest baseline is an MLP using these features.

\subsection{Results}
Our best performing LSTM (LSTM-DO-TR) used two layers of $128$ memory cells, dropout of probability 0.5 between layers, and target replication, and outperformed the MLP with hand-engineered features. Moreover simple ensembles of the best LSTM and MLP outperformed both on all metrics. \autoref{tab:experimental-results} shows summary results for all models. \autoref{tab:single-label-F1-results} shows the LSTM's predictive performance for six diagnoses with the highest F1 scores. Full per-diagnosis results can be found in \autoref{sec:perdiagnosis}.

\textit{Target replication} improves performance on all metrics, accelerating learning and reducing overfitting (\autoref{fig:overfitting}). 
We also find that the LSTM with target replication learns to output correct diagnoses earlier in the time series, a virtue that we explore qualitatively in \autoref{sec:hourly-diagnosis}. As a comparison, we trained a LSTM-DO-TR variant using the linear gain strategy of \cite{ng2015beyond,dai2015semi}. In general, this model did not perform as well as our simpler target replication strategy, but it did achieve the highest macro F1 score among the LSTM models.

\begin{table*}[hbt!]
\centering
\footnotesize
\begin{tabular}{llllll}
\multicolumn{6}{c}{\textbf{Classification performance for 128 ICU phenotypes}} \\
\toprule

\textbf{\textbf{Model}} & \textbf{Micro AUC} & \textbf{Macro AUC} & \textbf{Micro F1} & \textbf{Macro F1} & \textbf{Prec. at 10} \\
\midrule
\textbf{Base Rate} & 0.7128 & 0.5 & 0.1346 & 0.0343 & 0.0788 \\ % 0.2281
\textbf{Log. Reg., First 6 + Last 6} & 0.8122 & 0.7404 & 0.2324 & 0.1081 & 0.1016 \\
\textbf{Log. Reg., Expert features} & 0.8285 & 0.7644 & 0.2502 & 0.1373 & 0.1087 \\
\textbf{MLP, First 6 + Last 6} & 0.8375 & 0.7770 & 0.2698 & 0.1286 & 0.1096 \\
\textbf{MLP, Expert features} & \textbf{0.8551 } & \textbf{0.8030} & \textbf{0.2930} & \textbf{0.1475} & \textbf{0.1170} \\
\midrule
\multicolumn{6}{c}{\textbf{LSTM Models with two 64-cell hidden layers}} \\
\midrule
\textbf{LSTM} & 0.8241 & 0.7573 & 0.2450 & 0.1170 & 0.1047 \\
\textbf{LSTM, AuxOut (Diagnoses)} & 0.8351 & 0.7746 & 0.2627 & 0.1309 & 0.1110 \\
\textbf{LSTM-AO (Categories)} & 0.8382 & 0.7748 & 0.2651 & 0.1351 & 0.1099 \\
\textbf{LSTM-TR} & 0.8429 & 0.7870 & 0.2702 & 0.1348 & 0.1115 \\
\textbf{LSTM-TR-AO (Diagnoses)} & 0.8391 & 0.7866 & 0.2599 & 0.1317 & 0.1085 \\
\textbf{LSTM-TR-AO (Categories)} & 0.8439 & 0.7860 & 0.2774 & 0.1330 & 0.1138 \\
\midrule
 \multicolumn{6}{c}{\textbf{LSTM Models with Dropout (probability 0.5) and two 128-cell hidden layers}} \\
\midrule
\textbf{LSTM-DO} & 0.8377 & 0.7741 & 0.2748 & 0.1371 & 0.1110 \\
\textbf{LSTM-DO-AO (Diagnoses)} & 0.8365 & 0.7785 & 0.2581 & 0.1366 & 0.1104 \\
\textbf{LSTM-DO-AO (Categories)} & 0.8399 & 0.7783 & 0.2804 & 0.1361 & 0.1123 \\
\textbf{LSTM-DO-TR} & \textcolor{red}{\textbf{0.8560}} & \textcolor{red}{\textbf{0.8075}} & \textcolor{red}{\textbf{0.2938}} & 0.1485 & 0.\textcolor{red}{\textbf{1172}} \\
\textbf{LSTM-DO-TR-AO (Diagnoses)} & 0.8470 & 0.7929 & 0.2735 & 0.1488 & 0.1149 \\
\textbf{LSTM-DO-TR-AO (Categories)} & 0.8543 & 0.8015 & 0.2887 & 0.1446 & 0.1161 \\
\textbf{LSTM-DO-TR (Linear Gain)} & 0.8480 & 0.7986 & 0.2896 & \textcolor{red}{\textbf{0.1530}} & 0.1160 \\
\midrule
\multicolumn{6}{c}{\textbf{Ensembles of Best MLP and Best LSTM}} \\
\midrule
\textbf{Mean of LSTM-DO-TR \& MLP} & 0.8611 & 0.8143 & 0.2981 & 0.1553 & 0.1201  \\
\textbf{Max of LSTM-DO-TR \& MLP} & \textcolor{blue}{\textbf{0.8643}} & \textcolor{blue}{\textbf{0.8194}}  & \textcolor{blue}{\textbf{0.3035}} & \textcolor{blue}{\textbf{0.1571}} & \textcolor{blue}{\textbf{0.1218}} \\

\bottomrule
\end{tabular} \hspace{-0.2in}
\caption{Results on performance metrics calculated across all labels. \emph{DO}, \emph{TR}, and \emph{AO} indicate dropout, target replication, and \emph{auxiliary outputs}, respectively. \emph{AO (Diagnoses)} uses the extra diagnosis codes and \emph{AO (Categories)} uses diagnostic categories as additional targets during training.}
\label{tab:experimental-results}
\end{table*}

\emph{Auxiliary outputs} improved performance for most metrics 
and reduced overfitting. 
While the performance improvement is not as dramatic as that conferred by target replication, the regularizing effect is greater. 
These gains came at the cost of slower training: the auxiliary output models required more epochs (\autoref{fig:overfitting} and \autoref{sec:learning-curves}), 
especially when using the 301 remaining diagnoses. This may be due in part to severe class imbalance in the extra labels. 
For many of these labels it may take an entire epoch just to learn that they are occasionally nonzero.

\begin{table*}[hbt!]
\centering
\footnotesize
\begin{tabular}{lllll}
\multicolumn{5}{c}{\textbf{Top $6$ diagnoses measured by F1 score}} \\
\toprule
\textbf{Label} & \textbf{\textit{\textcolor{red}{F1}}} & \textbf{AUC} & \textbf{Precision} & \textbf{Recall} \\
\midrule
\textbf{Diabetes mellitus with ketoacidosis} & 0.8571 & 0.9966 & 1.0000 & 0.7500 \\
\textbf{Scoliosis, idiopathic} & 0.6809 & 0.8543 & 0.6957 & 0.6667 \\
\textbf{Asthma, unspecified with status asthmaticus} & 0.5641 & 0.9232 & 0.7857 & 0.4400 \\
\textbf{Neoplasm, brain, unspecified} & 0.5430 & 0.8522 & 0.4317 & 0.7315 \\
\textbf{Delayed milestones} & 0.4751 & 0.8178 & 0.4057 & 0.5733 \\
\textbf{Acute Respiratory Distress Syndrome (ARDS)} & 0.4688 & 0.9595 & 0.3409 & 0.7500 \\
\bottomrule
\end{tabular} \hspace{-0.2in}
\caption{LSTM-DO-TR performance on the $6$ diagnoses with highest F1 scores.}
\label{tab:single-label-F1-results}
\end{table*}

The LSTMs appear to learn models 
complementary to the MLP trained on hand-engineered features. 
Supporting this claim, simple ensembles of the LSTM-DO-TR and MLP (taking the \emph{mean} or \emph{maximum} of their predictions) outperform the constituent models significantly on all metrics (\autoref{tab:experimental-results}). Further, there are many diseases for which one model substantially outperforms the other, e.g., intracranial hypertension for the LSTM, septic shock for the MLP (\autoref{sec:perdiagnosis}).

\section{Discussion}
Our results indicate that LSTM RNNs, 
especially with target replication, 
can successfully classify diagnoses 
of critical care patients 
given clinical time series data.
The best LSTM beat a strong MLP baseline using hand-engineered features as input, and an ensemble combining the MLP and LSTM improves upon both.
The success of target replication accords with results by both \cite{ng2015beyond} and \cite{dai2015semi},
who observed similar benefits on their respective tasks. 
However, while they saw improvement using a linearly increasing weight on each target from start to end, 
this strategy performed worse in our diagnostic classification task than our uniform weighting of intermediate targets.
We believe this may owe to the peculiar nature of our data.
Linear gain emphasizes evidence from later in the sequence,
an assumption which often does not match the progression of symptoms in critical illnesses. 
Asthma patients, for example, are often admitted to the ICU severely symptomatic, 
but once treatment begins, patient physiology stabilizes and observable signs of disease may abate or change. 
Further supporting this idea, we observed that when training fixed-window baselines, 
using the first 6 and last 6 hours outperformed using the last 12 hours only.

\begin{wrapfigure}{r}{0.5\textwidth}
  \begin{center}
    \includegraphics[width=0.5\textwidth,clip=true,trim=55 10 70 50]{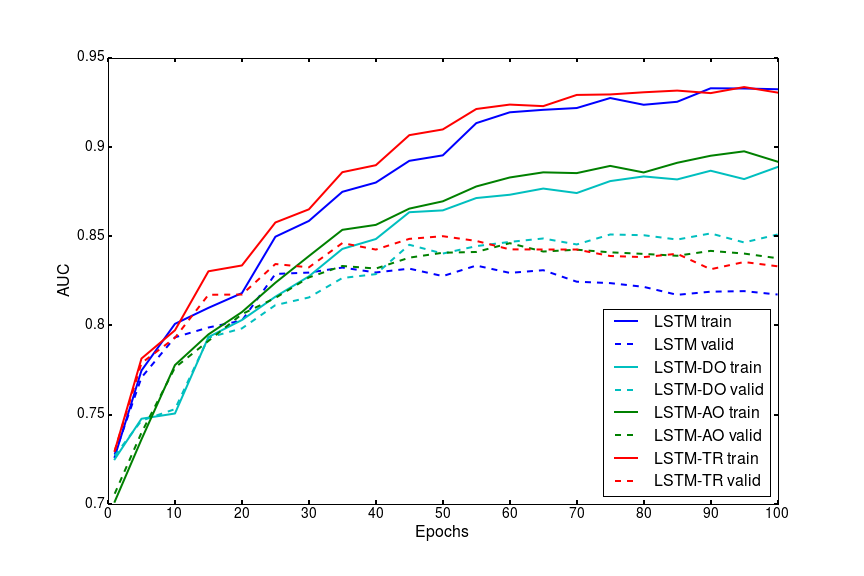}
  \end{center}
  \vspace{-15px}
  \caption{Training curves showing the impact of the \emph{DO}, \emph{AO}, and \emph{TR} strategies on overfitting.}
  \vspace{-10px}
  \label{fig:overfitting}
\end{wrapfigure}

While our data is of large scale by clinical standards,
it is small relative to datasets found in deep learning tasks like vision and speech recognition.
At this scale, regularization is critical.
Our experiments demonstrate that target replication, 
auxiliary outputs, and dropout 
all work to reduce the generalization gap. as shown in \autoref{fig:overfitting} and \autoref{sec:learning-curves}. 
However, some of these techniques are complementary 
while others seem to cancel each other out. 
For example, our best model combined target replication with dropout.
This combination significantly improved upon the performance using target replication alone,
and enabled the effective use of larger capacity models.
In contrast, the benefits of dropout and auxiliary output training appear to wash each other out. 
This may be because target replication confers more than regularization,
mitigating the difficulty of learning long range dependencies by providing local objectives.

\section{Conclusion}
While our LSTMs produce promising results, 
this is only a first step in this line of research.
Recognizing diagnoses given full time series of sensor data  
demonstrates that LSTMs can capture meaningful signal,
but ultimately we would like to predict developing conditions and events, outcomes such as mortality, and treatment responses.
In this paper we used diagnostic labels without timestamps, 
but we are obtaining timestamped diagnoses, 
which will enable us to train models to perform early diagnosis by predicting future conditions.
In addition, we are extending this work to a larger PICU data set
with 50\% more patients and hundreds of variables, including treatments and medications.

On the methodological side, 
we would like to both better exploit and improve the capabilities of LSTMs.
Results from speech recognition have shown 
that LSTMs shine in comparison to other models using raw features,
% In the clinical setting, LSTMs may allow us to exploit previously difficult to mine sources of data while
minimizing need for preprocessing and feature engineering.
In contrast, our current data preparation pipeline
removes valuable structure and information 
from clinical time series that could be exploited by an LSTM. 
For example, our forward- and back-filling imputation strategies 
discard useful information about when each observation is recorded. 
Imputing normal values for missing time series ignores the meaningful distinction between truly normal and missing measurements.
Also, our window-based resampling procedure reduces the variability of more frequently measured vital signs (e.g., heart rate). 

In future work, we plan to introduce indicator variables 
to allow the LSTM to distinguish actual from missing or imputed measurements. 
Additionally, the flexibility of the LSTM architecture 
should enable us to eliminate age-based corrections 
and to incorporate non-sequential inputs, such as age, weight, and height (or even hand-engineered features), into predictions.
Other next steps in this direction 
include developing LSTM architectures
to directly handle missing values and irregular sampling.
We also are encouraged by the success of target replication
and plan to explore other variants of this technique and to apply it to other domains and tasks.
Additionally, we acknowledge that there remains a debate 
about the interpretability of neural networks 
when applied to complex medical problems. 
We are developing methods to interpret 
the representations learned by LSTMs 
in order to better expose patterns of health and illness to clinical users.
We also hope to make practical use 
of the distributed representations of patients
for tasks such as patient similarity search.

\section{Acknowledgements}
Zachary C. Lipton was supported by the Division of Biomedical Informatics at the University of California, San Diego, via training grant (T15LM011271) from the NIH/NLM.
David Kale was supported by the Alfred E. Mann Innovation in Engineering Doctoral Fellowship.
% The authors would like to thank Randall Wetzel of Children's Hospital Los Angeles for his assistance in understanding and analyzing clinical data. 
The VPICU was supported by grants from the Laura P. and Leland K. Whittier Foundation.
We acknowledge NVIDIA Corporation for Tesla K40 GPU hardware donation and Professors Julian McAuley and Greg Ver Steeg for their support and advice. Finally, we thank the anonymous ICLR reviewers for their feedback, which helped us to make significant improvements to this work and manuscript.

\bibliography{iclr}
\bibliographystyle{iclr2016_conference}

\newpage

\appendix
\appendixpage
\section{Hourly Diagnostic Predictions}
\label{sec:hourly-diagnosis}
Our LSTM networks predict 128 diagnoses 
given sequences of clinical measurements. 
Because each network is connected left-to-right, i.e., in chronological order,
we can output predictions at each sequence step. 
Ultimately, we imagine that this capability 
could be used to make continuously updated real-time alerts and diagnoses.
Below, we explore this capability qualitatively. 
We choose examples of patients 
with a correctly classified diagnosis 
and visualize the probabilities assigned by each LSTM model at each sequence step. 
In addition to improving the quality 
of the final output, 
the LSTMs with target replication (LSTM-TR) 
arrive at correct diagnoses quickly 
compared to the simple multilabel LSTM model (LSTM-Simple). 
When auxiliary outputs are also used 
(LSTM-TR,AO), 
the diagnoses appear to be generally more confident.

\begin{figure}[ht]
\centering
\begin{subfigure}{.49\textwidth}
  \centering
  \includegraphics[width=.95\linewidth]{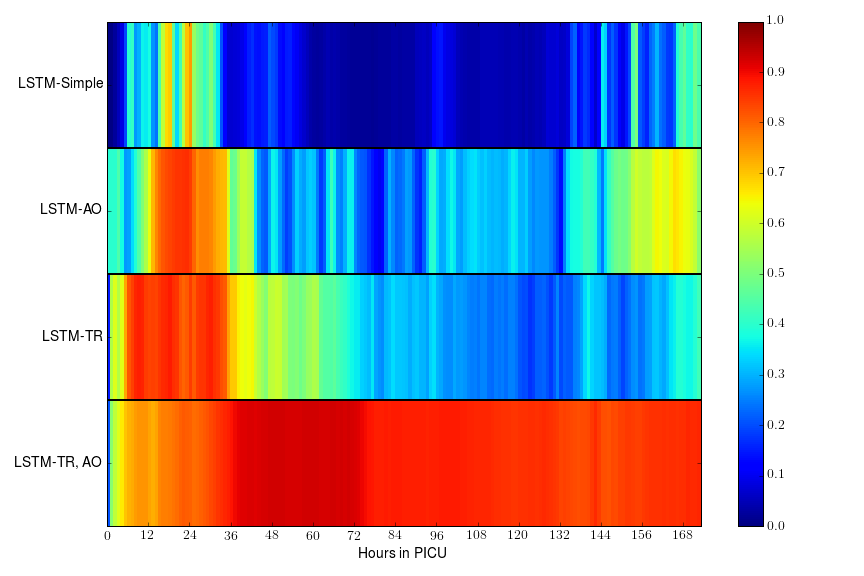}
  \caption{Asthma with Status Asthmaticus}
  \label{fig:asthma}
\end{subfigure}
\begin{subfigure}{.49\textwidth}
  \centering
  \includegraphics[width=.95\linewidth]{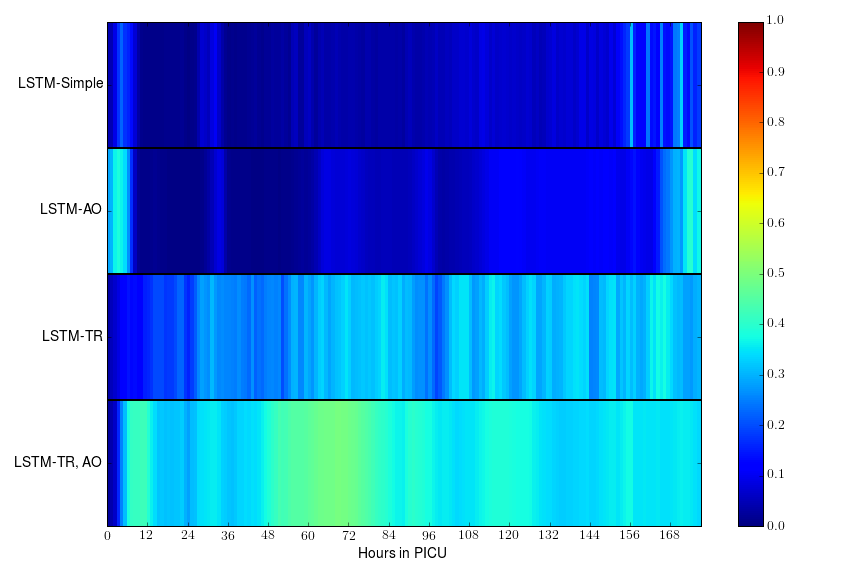}
  \caption{Acute Respiratory Distress Syndrome}
  \label{fig:ards}
\end{subfigure}
\begin{subfigure}{.49\textwidth}
  \centering
  \includegraphics[width=.95\linewidth]{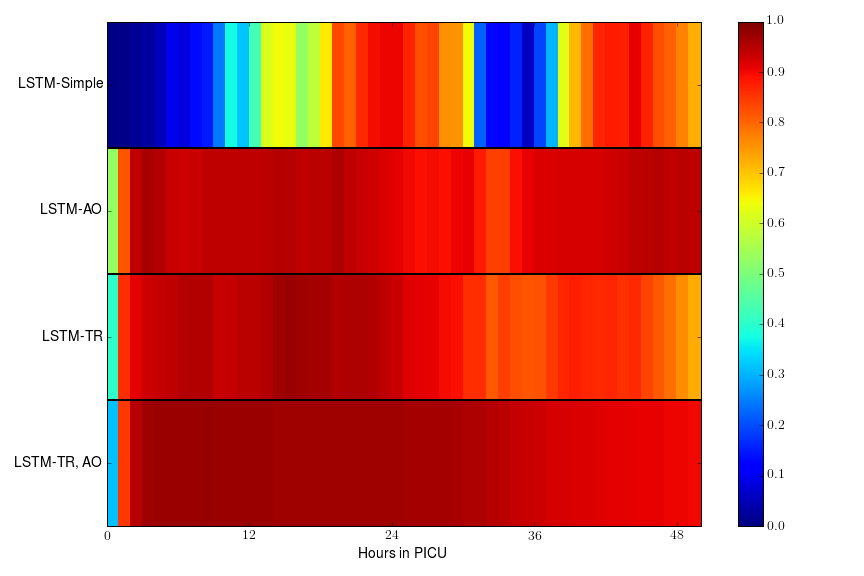}
  \caption{Diabetic Ketoacidosis}
  \label{fig:diabetes}
\end{subfigure}
\begin{subfigure}{.49\textwidth}
  \centering
  \includegraphics[width=.95\linewidth]{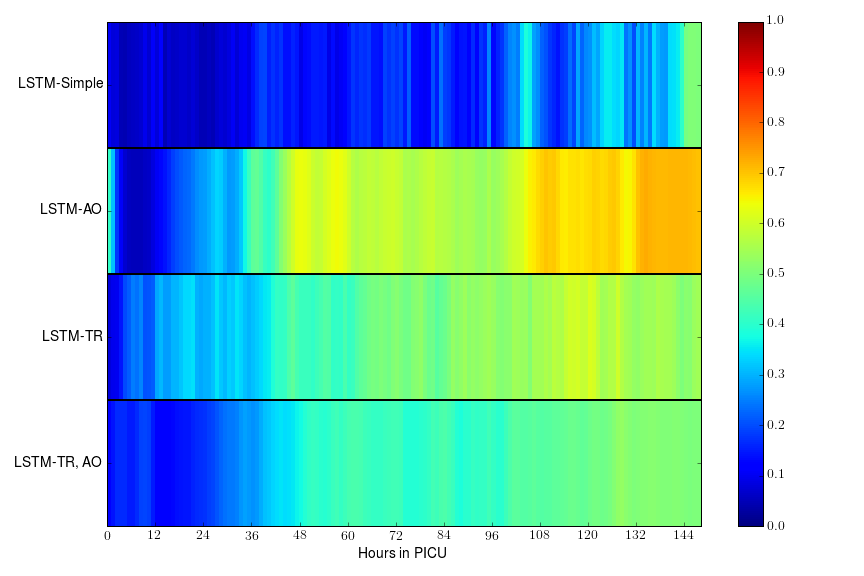}
  \caption{Brain Neoplasm, Unspecified Nature}
  \label{fig:neoplasm}
\end{subfigure}
\begin{subfigure}{.49\textwidth}
  \centering
  \includegraphics[width=.95\linewidth]{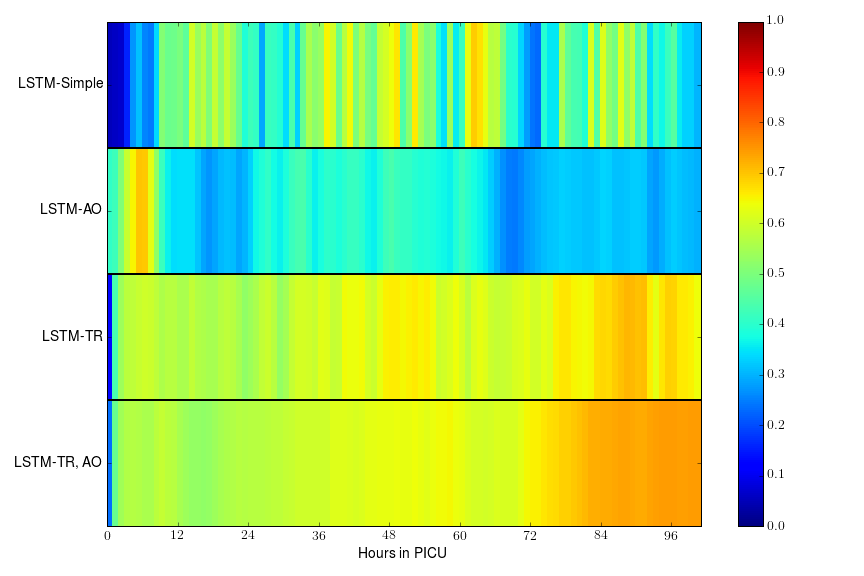}
  \caption{Septic Shock}
  \label{fig:shock}
\end{subfigure}
\begin{subfigure}{.49\textwidth}
  \centering
  \includegraphics[width=.95\linewidth]{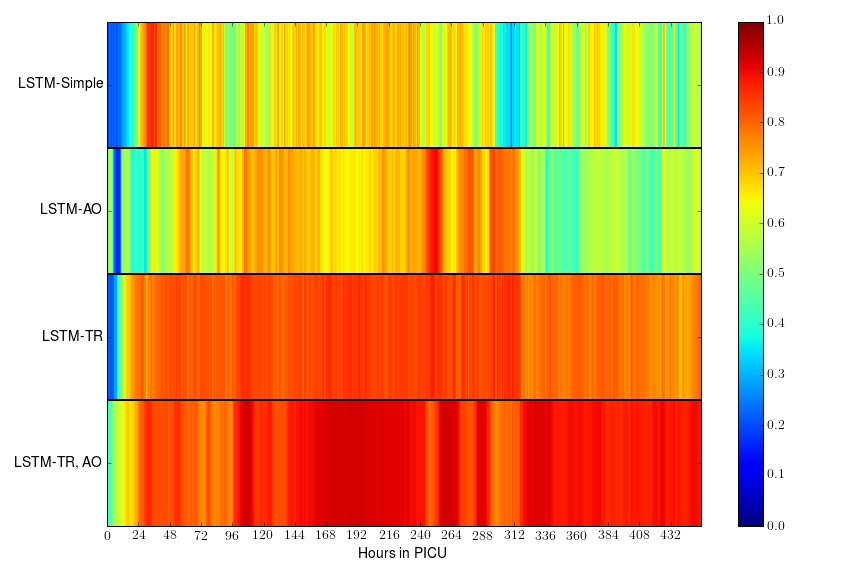}
  \caption{Scoliosis}
  \label{fig:Scoliosis}
\end{subfigure}
\caption{Each chart depicts the probabilities assigned by each of four models at each (hourly re-sampled) time step. LSTM-Simple uses only targets at the final time step. LSTM-TR uses target replication. LSTM-AO uses auxiliary outputs (diagnoses), and LSTM-TR,AO uses both techniques. LSTMs with target replication learn to make accurate diagnoses earlier.}
\label{fig:heatmaps}
\vspace{-10pt}
\end{figure} 
Our LSTM-TR,AO effectively predicts status asthmaticus and acute respiratory distress syndrome, 
likely owing to the several measures 
of pulmonary function among our inputs.
Diabetic ketoacidosis also proved easy to diagnose, 
likely because glucose and pH 
are included among our clinical measurements.
We were surprised to see that the network classified scoliosis reliably, 
but a deeper look into the medical literature suggests that scoliosis often results in respiratory symptoms. 
This analysis of step-by-step predictions is preliminary and informal, and we note that for a small number of examples our data preprocessing introduces a target leak by back-filling missing values.
In future work, when we explore this capability in greater depth, 
we will reprocess the data.

\section{Learning Curves}
\label{sec:learning-curves}
\begin{figure}[ht]
\centering
\begin{subfigure}{.49\textwidth}
  \centering
  \includegraphics[width=1.05\linewidth,clip=true,trim=60 0 50 0]{figures/overfitting-lineplot-AUC.png}
  \caption{AUC learning curves}
  \label{fig:auroc-curve}
\end{subfigure}
\begin{subfigure}{.49\textwidth}
  \centering
  \includegraphics[width=1.05\linewidth,clip=true,trim=60 0 50 0]{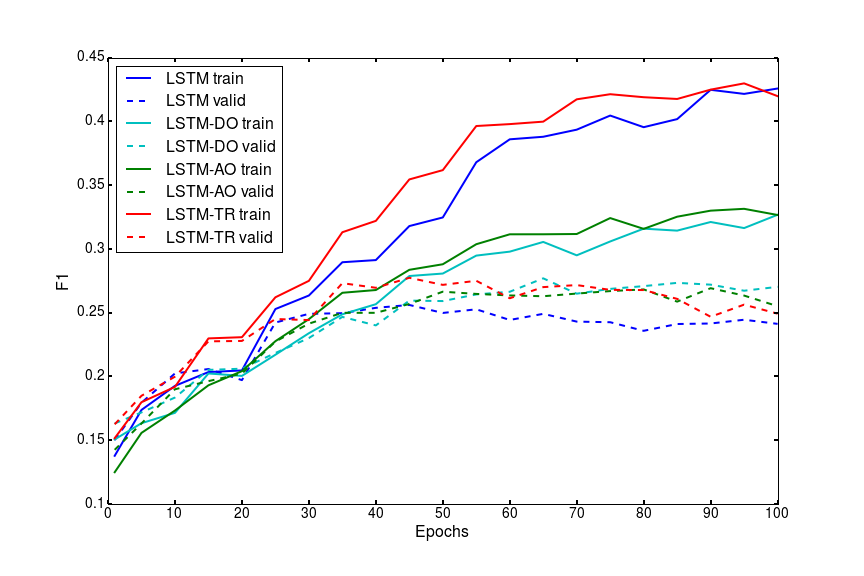}
  \caption{F1 learning curves}
  \label{fig:f1-curve}
\end{subfigure}
\begin{subfigure}{.49\textwidth}
  \centering
  \includegraphics[width=1.05\linewidth,clip=true,trim=60 0 50 0]{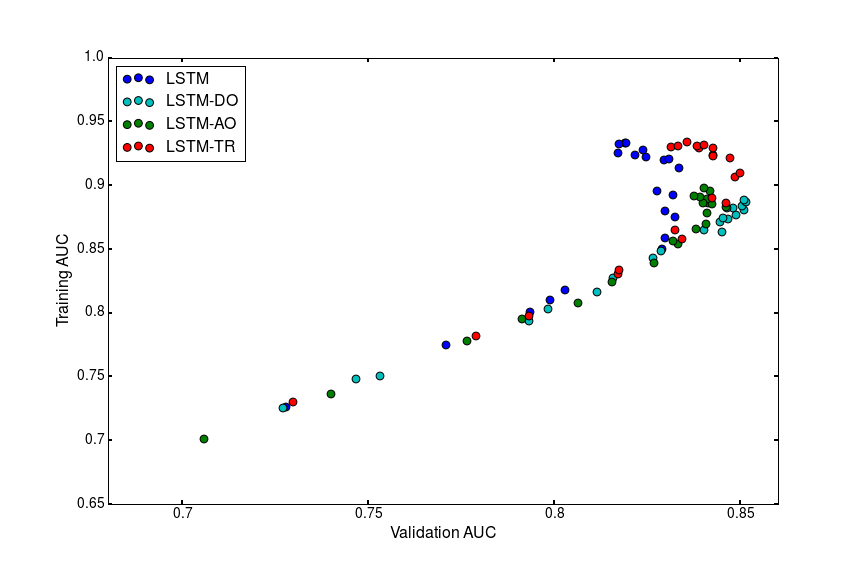}
  \caption{AUC training \textit{vs.} validation}
  \label{fig:auroc-scatter}
\end{subfigure}
\begin{subfigure}{.49\textwidth}
  \centering
  \includegraphics[width=1.05\linewidth,clip=true,trim=60 0 50 0]{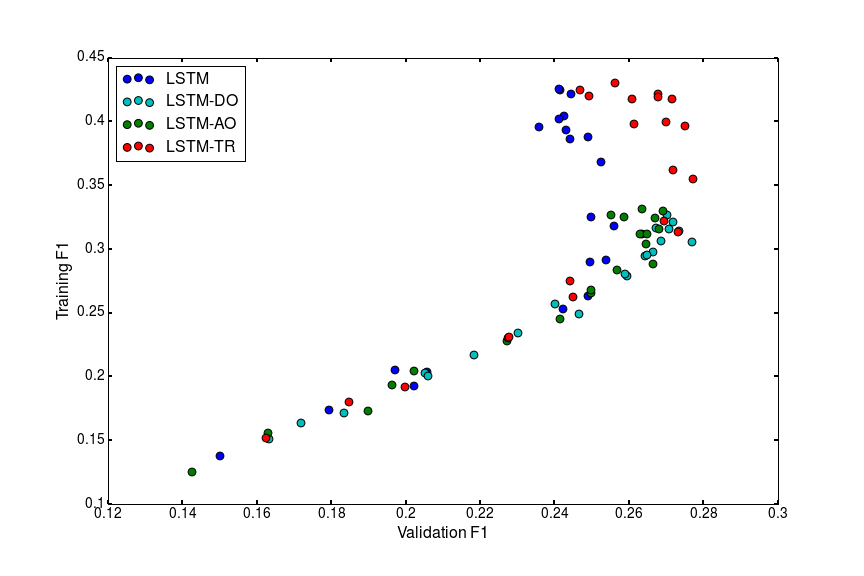}
  \caption{F1 training \textit{vs.} validation}
  \label{fig:f1-scatter}
\end{subfigure}
\caption{Training and validation performance plotted for the simple multilabel network (LSTM-Simple), LSTM with target replication (LSTM-TR), and LSTM with auxiliary outputs (LSTM-AO). Target replication appears to increase the speed of learning and confers a small regularizing effect. Auxiliary outputs slow down the speed of learning but impart a strong regularizing effect.}
\label{fig:learning-curves}
\vspace{-10pt}
\end{figure}

We present visualizations of the performance 
of LSTM, LSTM-DO (with dropout probability $0.5$), LSTM-AO (using the 301 additional diagnoses), and LSTM-TR (with $\alpha=0.5$), during training. 
These charts are useful for examining the effects 
of dropout, auxiliary outputs, and target replication
on both the speed of learning and the regularization they confer.
Specifically, for each of the four models, 
we plot the training and validation micro AUC and F1 score every five epochs in \autoref{fig:learning-curves}.
Additionally, we plot a scatter of the performance 
on the training set vs. the performance on the validation set.
The LSTM with target replication learns more quickly than a simple LSTM and also suffers less overfitting. 
With both dropout and auxiliary outputs, 
the LSTM trains more slowly than a simple LSTM 
but suffers considerably less overfitting.

\newpage
\section{Per Diagnosis Results}
\label{sec:perdiagnosis}
While averaged statistics provide an efficient way to check the relative quality of various models, 
considerable information is lost by reducing performance to a single scalar quantity. For some labels, our classifier makes classifications  with surprisingly high accuracy while for others, our features are uninformative and thus the classifier would not be practically useful.
To facilitate a more granular investigation of our model's predictive power, we present individual test set F1 and AUC scores for each individual diagnostic label in \autoref{tab:individual-results}. We compare the performance our best LSTM, which combines two 128-cell hidden layers with \emph{dropout} of probability 0.5 and \emph{target replication}, against the strongest baseline, an MLP trained on the hand-engineered features, and an ensemble predicts the maximum probability of the two. The results are sorted in descending order using the F1 performance of the LSTM, providing insights into the types of conditions that the LSTM can successfully classify.

\begin{table*}[ht]
\centering
\scriptsize
\begin{tabular}{lcccccc}
\multicolumn{7}{c}{\textbf{Classifier Performance on Each Diagnostic Code, Sorted by F1}} \\
\toprule
 & \multicolumn{2}{c}{\textbf{LSTM-DO-TR}} & \multicolumn{2}{c}{\textbf{MLP, Expert features}} & \multicolumn{2}{c}{\textbf{Max Ensemble}} \\
\textbf{Condition} & \textit{\textcolor{red}{F1}} & AUC & F1 & AUC & F1 & AUC \\
\midrule

\textbf{Diabetes mellitus with ketoacidosis} & 0.8571 & 0.9966 & 0.8571 & 0.9966 & 0.8571 & 0.9966 \\
\textbf{Scoliosis, idiopathic} & 0.6809 & 0.8543 & 0.6169 & 0.8467 & 0.6689 & 0.8591 \\
\textbf{Asthma, unspecified with status asthmaticus} & 0.5641 & 0.9232 & 0.6296 & 0.9544 & 0.6667 & 0.9490 \\
\textbf{Neoplasm, brain, unspecified nature} & 0.5430 & 0.8522 & 0.5263 & 0.8463 & 0.5616 & 0.8618 \\
\textbf{Developmental delay} & 0.4751 & 0.8178 & 0.4023 & 0.8294 & 0.4434 & 0.8344 \\
\textbf{Acute respiratory distress syndrome (ARDS)} & 0.4688 & 0.9595 & 0.3913 & 0.9645 & 0.4211 & 0.9650 \\
\textbf{Hypertension, unspecified} & 0.4118 & 0.8593 & 0.3704 & 0.8637 & 0.3636 & 0.8652 \\
\textbf{Arteriovenous malformation of brain} & 0.4000 & 0.8620 & 0.3750 & 0.8633 & 0.3600 & 0.8684 \\
\textbf{End stage renal disease on dialysis} & 0.3889 & 0.8436 & 0.3810 & 0.8419 & 0.3902 & 0.8464 \\
\textbf{Acute respiratory failure} & 0.3864 & 0.7960 & 0.4128 & 0.7990 & 0.4155 & 0.8016 \\
\textbf{Renal transplant status post} & 0.3846 & 0.9692 & 0.4828 & 0.9693 & 0.4800 & 0.9713 \\
\textbf{Epilepsy, unspecified, not intractable} & 0.3740 & 0.7577 & 0.3145 & 0.7265 & 0.3795 & 0.7477 \\
\textbf{Septic shock} & 0.3721 & 0.8182 & 0.3210 & 0.8640 & 0.3519 & 0.8546 \\
\textbf{Other respiratory symptom} & 0.3690 & 0.8088 & 0.3642 & 0.7898 & 0.3955 & 0.8114 \\
\textbf{Biliary atresia} & 0.3636 & 0.9528 & 0.5000 & 0.9338 & 0.4444 & 0.9541 \\
\textbf{Acute lymphoid leukemia, without remission} & 0.3486 & 0.8601 & 0.3288 & 0.8293 & 0.3175 & 0.8441 \\
\textbf{Congenital hereditary muscular dystrophy} & 0.3478 & 0.8233 & 0.0000 &  0.8337 & 0.2727 & 0.8778 \\
\textbf{Liver transplant status post} & 0.3448 & 0.8431 & 0.3333 & 0.8104 & 0.3846 & 0.8349 \\
\textbf{Respiratory complications, prodecure status post} & 0.3143 & 0.8545 & 0.2133 & 0.8614 & 0.3438 & 0.8672 \\
\textbf{Grand mal status} & 0.3067 & 0.8003 & 0.3883 & 0.7917 & 0.3529 & 0.8088 \\
\textbf{Intracranial injury, closed} & 0.3048 & 0.8589 & 0.3095 & 0.8621 & 0.3297 & 0.8820 \\
\textbf{Diabetes insipidus} & 0.2963 & 0.9455 & 0.3774 & 0.9372 & 0.4068 & 0.9578 \\
\textbf{Acute renal failure, unspecified} & 0.2553 & 0.8806 & 0.2472 & 0.8698 & 0.2951 & 0.8821 \\
\textbf{Other diseases of the respiratory system} & 0.2529 & 0.7999 & 0.1864 & 0.7920 & 0.2400 & 0.8131 \\
\textbf{Croup syndrome} & 0.2500 & 0.9171 & 0.1538 & 0.9183 & 0.0000 &  0.9263 \\
\textbf{Bronchiolitis due to other infectious organism} & 0.2466 & 0.9386 & 0.2353 & 0.9315 & 0.2712 & 0.9425 \\
\textbf{Congestive heart failure} & 0.2439 & 0.8857 & 0.0000 &  0.8797 & 0.0000 &  0.8872 \\
\textbf{Infantile cerebral palsy, unspecified} & 0.2400 & 0.8538 & 0.1569 & 0.8492 & 0.2083 & 0.8515 \\
\textbf{Congenital hydrocephalus} & 0.2393 & 0.7280 & 0.2247 & 0.7337 & 0.1875 & 0.7444 \\
\textbf{Cerebral edema} & 0.2222 & 0.8823 & 0.2105 & 0.9143 & 0.2500 & 0.9190 \\
\textbf{Craniosynostosis} & 0.2222 & 0.8305 & 0.5333 & 0.8521 & 0.6154 & 0.8658 \\
\textbf{Anoxic brain damage} & 0.2222 & 0.8108 & 0.1333 & 0.8134 & 0.2500 & 0.8193 \\
\textbf{Pneumonitis due to inhalation of food or vomitus} & 0.2222 & 0.6547 & 0.0326 & 0.6776 & 0.0462 & 0.6905 \\
\textbf{Acute and subacute necrosis of the liver} & 0.2182 & 0.8674 & 0.2778 & 0.9039 & 0.2381 & 0.8964 \\
\textbf{Respiratory syncytial virus} & 0.2154 & 0.9118 & 0.1143 & 0.8694 & 0.1622 & 0.9031 \\
\textbf{Unspecified disorder of kidney and ureter} & 0.2069 & 0.8367 & 0.1667 & 0.8496 & 0.1667 & 0.8559 \\
\textbf{Craniofacial malformation} & 0.2059 & 0.8688 & 0.4444 & 0.8633 & 0.3158 & 0.8866 \\
\textbf{Pulmonary hypertension, secondary} & 0.2000 & 0.9377 & 0.0870 & 0.8969 & 0.2105 & 0.9343 \\
\textbf{Bronchopulmonary dysplasia} & 0.1905 & 0.8427 & 0.1404 & 0.8438 & 0.1333 & 0.8617 \\
\textbf{Drowning and non-fatal submersion} & 0.1905 & 0.8341 & 0.1538 & 0.8905 & 0.1429 & 0.8792 \\
\textbf{Genetic abnormality} & 0.1828 & 0.6727 & 0.1077 & 0.6343 & 0.1111 & 0.6745 \\
\textbf{Other and unspecified coagulation defects} & 0.1818 & 0.7081 & 0.0000 &  0.7507 & 0.1600 & 0.7328 \\
\textbf{Vehicular trauma} & 0.1778 & 0.8655 & 0.2642 & 0.8505 & 0.2295 & 0.8723 \\

\end{tabular} \hspace{-0.2in}
% \caption{Results on performance metrics calculated on the $5$ labels with highest F1.}
\caption{F1 and AUC scores for individual diagnoses.}
\label{tab:individual-results}
\end{table*}

\begin{table*}[ht]
\centering
\scriptsize
\begin{tabular}{lcccccc}
\multicolumn{7}{c}{\textbf{Classifier Performance on Each Diagnostic Code, Sorted by F1}} \\
\toprule
 & \multicolumn{2}{c}{\textbf{LSTM-DO-TR}} & \multicolumn{2}{c}{\textbf{MLP, Expert features}} & \multicolumn{2}{c}{\textbf{Max Ensemble}} \\
\textbf{Condition} & \textit{\textcolor{red}{F1}} & AUC & F1 & AUC & F1 & AUC \\
\midrule

\textbf{Other specified cardiac dysrhythmia} & 0.1667 & 0.7698 & 0.1250 & 0.8411 & 0.0800 & 0.8179 \\
\textbf{Acute pancreatitis} & 0.1622 & 0.8286 & 0.1053 & 0.8087 & 0.1379 & 0.8440 \\
\textbf{Esophageal reflux} & 0.1515 & 0.8236 & 0.0000 &  0.7774 & 0.1739 & 0.8090 \\
\textbf{Cardiac arrest, outside hospital} & 0.1500 & 0.8562 & 0.1333 & 0.9004 & 0.1765 & 0.8964 \\
\textbf{Unspecified pleural effusion} & 0.1458 & 0.8777 & 0.1194 & 0.8190 & 0.1250 & 0.8656 \\
\textbf{Mycoplasma pneumoniae} & 0.1429 & 0.8978 & 0.1067 & 0.8852 & 0.1505 & 0.8955 \\
\textbf{Unspecified immunologic disorder} & 0.1429 & 0.8481 & 0.1000 & 0.8692 & 0.1111 & 0.8692 \\
\textbf{Congenital alveolar hypoventilation syndrome} & 0.1429 & 0.6381 & 0.0000 &  0.7609 & 0.0000 &  0.7246 \\
\textbf{Septicemia, unspecified} & 0.1395 & 0.8595 & 0.1695 & 0.8640 & 0.1905 & 0.8663 \\
\textbf{Pneumonia due to adenovirus} & 0.1379 & 0.8467 & 0.0690 & 0.9121 & 0.1277 & 0.8947 \\
\textbf{Insomnia with sleep apnea} & 0.1359 & 0.7892 & 0.0752 & 0.7211 & 0.0899 & 0.8089 \\
\textbf{Defibrination syndrome} & 0.1333 & 0.9339 & 0.1935 & 0.9461 & 0.2500 & 0.9460 \\
\textbf{Unspecified injury, unspecified site} & 0.1333 & 0.8749 & 0.0000 &  0.7673 & 0.1250 & 0.8314 \\
\textbf{Pneumococcal pneumonia} & 0.1290 & 0.8706 & 0.1149 & 0.8664 & 0.1461 & 0.8727 \\
\textbf{Genetic or other unspecified anomaly} & 0.1277 & 0.7830 & 0.0870 & 0.7812 & 0.1429 & 0.7905 \\
\textbf{Other spontaneous pneumothorax} & 0.1212 & 0.8029 & 0.0972 & 0.8058 & 0.1156 & 0.8122 \\
\textbf{Bone marrow transplant status} & 0.1176 & 0.8136 & 0.0000 &  0.8854 & 0.2353 & 0.8638 \\
\textbf{Other primary cardiomyopathies} & 0.1176 & 0.6862 & 0.0000 &  0.6371 & 0.1212 & 0.6635 \\
\textbf{Intracranial hemorrhage} & 0.1071 & 0.7498 & 0.1458 & 0.7306 & 0.1587 & 0.7540 \\
\textbf{Benign intracranial hypertension} & 0.1053 & 0.9118 & 0.0909 & 0.7613 & 0.1379 & 0.8829 \\
\textbf{Encephalopathy, unspecified} & 0.1053 & 0.8466 & 0.0909 & 0.7886 & 0.0000 &  0.8300 \\
\textbf{Ventricular septal defect} & 0.1053 & 0.6781 & 0.0741 & 0.6534 & 0.0833 & 0.6667 \\
\textbf{Crushing injury, unspecified} & 0.1017 & 0.9183 & 0.0952 & 0.8742 & 0.1200 & 0.9111 \\
\textbf{Malignant neoplasm, disseminated} & 0.0984 & 0.7639 & 0.0588 & 0.7635 & 0.0667 & 0.7812 \\
\textbf{Orthopaedic surgery, post status} & 0.0976 & 0.7605 & 0.1290 & 0.8234 & 0.0845 & 0.8106 \\
\textbf{Thoracic surgery, post status} & 0.0930 & 0.9160 & 0.0432 & 0.7401 & 0.0463 & 0.9137 \\
\textbf{Ostium secundum type atrial septal defect} & 0.0923 & 0.7876 & 0.1538 & 0.8068 & 0.1154 & 0.7998 \\
\textbf{Malignant neoplasm, in gastrointestinal organs} & 0.0853 & 0.8067 & 0.1111 & 0.7226 & 0.1412 & 0.7991 \\
\textbf{Coma} & 0.0833 & 0.7255 & 0.1111 & 0.6542 & 0.1250 & 0.7224 \\
\textbf{Pneumonia due to inhalation of food or vomitus} & 0.0800 & 0.8282 & 0.0923 & 0.8090 & 0.0952 & 0.8422 \\
\textbf{Extradural hemorrage from injury, no open wound} & 0.0769 & 0.7829 & 0.0000 &  0.8339 & 0.0988 & 0.8246 \\
\textbf{Prematurity (less than 37 weeks gestation)} & 0.0759 & 0.7542 & 0.1628 & 0.7345 & 0.1316 & 0.7530 \\
\textbf{Asthma, unspecified, without status asthmaticus} & 0.0734 & 0.6679 & 0.0784 & 0.6914 & 0.0678 & 0.6867 \\
\textbf{Gastrointestinal surgery, post status} & 0.0714 & 0.7183 & 0.0984 & 0.6999 & 0.0851 & 0.7069 \\
\textbf{Nervous disorder, not elsewhere classified} & 0.0708 & 0.7127 & 0.1374 & 0.7589 & 0.1404 & 0.7429 \\
\textbf{Unspecified gastrointestinal disorder} & 0.0702 & 0.6372 & 0.0348 & 0.6831 & 0.0317 & 0.6713 \\
\textbf{Pulmonary congestion and hypostasis} & 0.0678 & 0.8359 & 0.0000 &  0.8633 & 0.0000 &  0.8687 \\
\textbf{Thrombocytopenia, unspecified} & 0.0660 & 0.7652 & 0.0000 &  0.7185 & 0.0000 &  0.7360 \\
\textbf{Lung contusion, no open wound} & 0.0639 & 0.9237 & 0.0000 &  0.9129 & 0.2222 & 0.9359 \\
\textbf{Acute pericarditis, unspecified} & 0.0625 & 0.8601 & 0.0000 &  0.9132 & 0.0000 &  0.9089 \\
\textbf{Nervous system complications from implant} & 0.0597 & 0.6727 & 0.0368 & 0.7082 & 0.0419 & 0.7129 \\
\textbf{Heart disease, unspecified} & 0.0588 & 0.8372 & 0.0000 &  0.8020 & 0.0000 &  0.8264 \\
\textbf{Suspected infection in newborn or infant} & 0.0588 & 0.6593 & 0.0000 &  0.7090 & 0.0606 & 0.6954 \\

\end{tabular} \hspace{-0.2in}
% \caption{F1 and AUC scores for individual diagnoses.}
% \label{tab:individual-results-2}

\end{table*}

\begin{table*}[ht]
\centering
\scriptsize
\begin{tabular}{lcccccc}
\multicolumn{7}{c}{\textbf{Classifier Performance on Each Diagnostic Code, Sorted by F1}} \\
\toprule
 & \multicolumn{2}{c}{\textbf{LSTM-DO-TR}} & \multicolumn{2}{c}{\textbf{MLP, Expert features}} & \multicolumn{2}{c}{\textbf{Max Ensemble}} \\
\textbf{Condition} & \textit{\textcolor{red}{F1}} & AUC & F1 & AUC & F1 & AUC \\
\midrule

\textbf{Anemia, unspecified} & 0.0541 & 0.7782 & 0.0488 & 0.7019 & 0.0727 & 0.7380 \\
\textbf{Muscular disorder, not elsewhere classified} & 0.0536 & 0.6996 & 0.0000 &  0.7354 & 0.1000 & 0.7276 \\
\textbf{Malignant neoplasm, adrenal gland} & 0.0472 & 0.6960 & 0.0727 & 0.6682 & 0.0548 & 0.6846 \\
\textbf{Hematologic disorder, unspecified} & 0.0465 & 0.7315 & 0.1194 & 0.7404 & 0.0714 & 0.7446 \\
\textbf{Hematemesis} & 0.0455 & 0.8116 & 0.0674 & 0.7887 & 0.0588 & 0.8103 \\
\textbf{Dehydration} & 0.0435 & 0.7317 & 0.1739 & 0.7287 & 0.0870 & 0.7552 \\
\textbf{Unspecified disease of spinal cord} & 0.0432 & 0.7153 & 0.0571 & 0.7481 & 0.0537 & 0.7388 \\
\textbf{Neurofibromatosis, unspecified} & 0.0403 & 0.7494 & 0.0516 & 0.7458 & 0.0613 & 0.7671 \\
\textbf{Intra-abdominal injury, no open wound} & 0.0333 & 0.7682 & 0.1569 & 0.8602 & 0.0690 & 0.8220 \\
\textbf{Thyroid disorder, unspecified} & 0.0293 & 0.5969 & 0.0548 & 0.5653 & 0.0336 & 0.6062 \\
\textbf{Hereditary hemolytic anemia, unspecifed} & 0.0290 & 0.7474 & 0.0000 &  0.6182 & 0.0000 &  0.6962 \\
\textbf{Subdural hemorrage, no open wound} & 0.0263 & 0.7620 & 0.1132 & 0.7353 & 0.0444 & 0.7731 \\
\textbf{Unspecified intestinal obstruction} & 0.0260 & 0.6210 & 0.2041 & 0.7684 & 0.0606 & 0.7277 \\
\textbf{Hyposmolality and/or hyponatremia} & 0.0234 & 0.6999 & 0.0000 &  0.7565 & 0.0000 &  0.7502 \\
\textbf{Primary malignant neoplasm, thorax} & 0.0233 & 0.6154 & 0.0364 & 0.6086 & 0.0323 & 0.5996 \\
\textbf{Supraventricular premature beats} & 0.0185 & 0.8278 & 0.0190 & 0.7577 & 0.0299 & 0.8146 \\
\textbf{Injury to intrathoracic organs, no open wound} & 0.0115 & 0.8354 & 0.0000 &  0.8681 & 0.0000 &  0.8604 \\
\textbf{Child abuse, unspecified} & 0.0000 &  0.9273 & 0.3158 & 0.9417 & 0.1818 & 0.9406 \\
\textbf{Acidosis} & 0.0000 &  0.9191 & 0.1176 & 0.9260 & 0.0000 &  0.9306 \\
\textbf{Infantile spinal muscular atrophy} & 0.0000 &  0.9158 & 0.0000 &  0.8511 & 0.0000 &  0.9641 \\
\textbf{Fracture, femoral shaft} & 0.0000 &  0.9116 & 0.0000 &  0.9372 & 0.0513 & 0.9233 \\
\textbf{Cystic fibrosis with pulmonary manifestations} & 0.0000 &  0.8927 & 0.0000 &  0.8086 & 0.0571 & 0.8852 \\
\textbf{Panhypopituitarism} & 0.0000 &  0.8799 & 0.2222 & 0.8799 & 0.0500 & 0.8872 \\
\textbf{Blood in stool} & 0.0000 &  0.8424 & 0.0000 &  0.8443 & 0.0000 &  0.8872 \\
\textbf{Sickle-cell anemia, unspecified} & 0.0000 &  0.8268 & 0.0000 &  0.7317 & 0.0000 &  0.7867 \\
\textbf{Cardiac dysrhythmia, unspecified} & 0.0000 &  0.8202 & 0.0702 & 0.8372 & 0.0000 &  0.8523 \\
\textbf{Agranulocytosis} & 0.0000 &  0.8157 & 0.1818 & 0.8011 & 0.1667 & 0.8028 \\
\textbf{Malignancy of bone, no site specified} & 0.0000 &  0.8128 & 0.0870 & 0.7763 & 0.0667 & 0.8318 \\
\textbf{Pneumonia, organism unspecified} & 0.0000 &  0.8008 & 0.0952 & 0.8146 & 0.0000 &  0.8171 \\
\textbf{Unspecified metabolic disorder} & 0.0000 &  0.7914 & 0.0000 &  0.6719 & 0.0000 &  0.7283 \\
\textbf{Urinary tract infection, no site specified} & 0.0000 &  0.7867 & 0.0840 & 0.7719 & 0.2286 & 0.7890 \\
\textbf{Obesity, unspecified} & 0.0000 &  0.7826 & 0.0556 & 0.7550 & 0.0000 &  0.7872 \\
\textbf{Apnea} & 0.0000 &  0.7822 & 0.2703 & 0.8189 & 0.0000 &  0.8083 \\
\textbf{Respiratory arrest} & 0.0000 &  0.7729 & 0.0000 &  0.8592 & 0.0000 &  0.8346 \\
\textbf{Hypovolemic shock} & 0.0000 &  0.7686 & 0.0000 &  0.8293 & 0.0000 &  0.8296 \\
\textbf{Hemophilus meningitis} & 0.0000 &  0.7649 & 0.0000 &  0.7877 & 0.0000 &  0.7721 \\
\textbf{Diabetes mellitus, type I, stable} & 0.0000 &  0.7329 & 0.0667 & 0.7435 & 0.0833 & 0.7410 \\
\textbf{Tetralogy of fallot} & 0.0000 &  0.7326 & 0.0000 &  0.6134 & 0.0000 &  0.6738 \\
\textbf{Congenital heart disease, unspecified} & 0.0000 &  0.7270 & 0.1333 & 0.7251 & 0.0000 &  0.7319 \\
\textbf{Mechanical complication of V-P shunt} & 0.0000 &  0.7173 & 0.0000 &  0.7308 & 0.0000 &  0.7205 \\
\textbf{Respiratory complications due to procedure} & 0.0000 &  0.7024 & 0.0000 &  0.7244 & 0.0000 &  0.7323 \\
\textbf{Teenage cerebral artery occlusion and infarction} & 0.0000 &  0.6377 & 0.0000 &  0.5982 & 0.0000 &  0.6507 \\

\bottomrule
\end{tabular} \hspace{-0.2in}
% \caption{F1 and AUC scores for individual diagnoses.}
% \label{tab:individual-results-3}

\end{table*}

\end{document}